\documentclass[sigconf,final]{acmart}

\AtBeginDocument{%
  \providecommand\BibTeX{{%
    \normalfont B\kern-0.5em{\scshape i\kern-0.25em b}\kern-0.8em\TeX}}}

\setcopyright{acmlicensed}
\copyrightyear{2024}
\acmYear{2024}
\acmDOI{XXXXXXX.XXXXXXX}
\acmConference[MM'24]{Make sure to enter the correct
  conference title from your rights confirmation emai}{June 03--05,
  2018}{Woodstock, NY}
\acmISBN{978-1-4503-XXXX-X/18/06}

\acmSubmissionID{4415}

\usepackage{multirow}
\usepackage{enumitem}
\usepackage{algorithm,amsmath,algorithmic,bm}
\begin{document}

\title{Core Knowledge Learning Framework for Graph Adaptation and Scalability Learning}


\author{Bowen Zhang, Zhichao Huang, Genan Dai, Guangning Xu, Xiaomao Fan, Hu Huang}


\begin{abstract}
  Graph classification is a pivotal challenge in machine learning, especially within the realm of graph-based data, given its importance in numerous real-world applications such as social network analysis, recommendation systems, and bioinformatics. Despite its significance, graph classification faces several hurdles, including adapting to diverse prediction tasks, training across multiple target domains, and handling small-sample prediction scenarios. Current methods often tackle these challenges individually, leading to fragmented solutions that lack a holistic approach to the overarching problem. In this paper, we propose an algorithm aimed at addressing the aforementioned challenges. By incorporating insights from various types of tasks, our method aims to enhance adaptability, scalability, and generalizability in graph classification. Motivated by the recognition that the underlying subgraph plays a crucial role in GNN prediction, while the remainder is task-irrelevant, we introduce the Core Knowledge Learning (\method{}) framework for graph adaptation and scalability learning. \method{} comprises several key modules, including the core subgraph knowledge submodule, graph domain adaptation module, and few-shot learning module for downstream tasks. Each module is tailored to tackle specific challenges in graph classification, such as domain shift, label inconsistencies, and data scarcity. By learning the core subgraph of the entire graph, we focus on the most pertinent features for task relevance. Consequently, our method offers benefits such as improved model performance, increased domain adaptability, and enhanced robustness to domain variations. Experimental results demonstrate significant performance enhancements achieved by our method compared to state-of-the-art approaches. Specifically, our method achieves notable improvements in accuracy and generalization across various datasets and evaluation metrics, underscoring its effectiveness in addressing the challenges of graph classification.
\end{abstract}



\def\method{CKL}


\maketitle

\section{Introduction}

Graphs have garnered considerable attention for their ability to represent structured and relational data across diverse fields, as noted in several studies~\cite{liu2021self,dai2021hyperbolic,xu2021discrimination,wang2021visual,peng2021attention}. Graph classification, a fundamental aspect of data analysis, focuses on predicting whole graph properties and has seen substantial research activity in recent years~\cite{yin2024dynamic,wang2021curgraph,ma2020adaptive,yin2024continuous}. This research has practical implications in various applications such as determining the quantum mechanical properties of molecules, including mutagenicity and toxicity~\cite{hao2020asgn}, and identifying the functions of chemical compounds~\cite{kojima2020kgcn}. 
A wide array of graph classification methodologies have been developed, with the majority leveraging Graph Neural Networks (GNNs) to deliver strong performance~\cite{xu2019powerful,kipf2017semi,yin2022dynamic,yin2023messages}. These methods typically utilize a neighbor-aware message passing mechanism coupled with a readout function to learn discriminative graph representations that effectively reflect the structural topology, thereby facilitating accurate classification.

Despite its considerable potential, graph classification faces several significant challenges that hinder its broader adoption and effectiveness. These challenges can be broadly categorized into three main areas:
(1) \textit{Label Aspect}: Graph classification models are often designed for specific tasks, which limits their ability to transfer knowledge to different prediction tasks. This lack of task-agnostic adaptability reduces the models' versatility and applicability across various domains~\cite{ju2024survey}. Additionally, differences in labeling standards or the quality of annotations across domains can lead to inconsistencies in model predictions, thus affecting overall performance and generalizability.
(2) \textit{Domain Shift Aspect}: Graph classification models are generally trained on a single target domain, which diminishes their effectiveness when applied to diverse domains. Adapting these models to various target domains is a significant challenge due to variations in data distribution~\cite{yin2023dream,pang2023sa,yin2023coco}, which can degrade performance. Domain shifts, marked by changes in the data distribution between the source and target domains, amplify this challenge. Effective adaptation mechanisms are essential to maintain model robustness and enhance generalization capabilities.
(3) \textit{Data Aspect}: Graph classification struggles with effectively handling small-sample prediction scenarios. The lack of sufficient labeled data in the source domain, combined with data scarcity in the target domain, presents considerable challenges to the adaptation process, potentially leading to poor generalization performance~\cite{finn2017model,liu2018learning,kim2019edge,altae2017low,wang2021property}. Additionally, imbalanced data distributions between domains compound these difficulties, calling for strategies to alleviate the effects of data scarcity and ensure fair and accurate model predictions across different domains.

In this paper, we present a new framework meticulously crafted to address the shortcomings of existing graph classification methods. Inspired by \cite{luo2020parameterized}, the framework identifies and separates the essential underlying subgraph that significantly impacts GNN predictions from the task-irrelevant portions of the graph. Our approach leverages fundamental principles of graph-based learning to effectively address these issues.
As depicted in Fig.~\ref{fig1}, we utilize \method{} to efficiently extract the core subgraph, which is then used for downstream applications such as graph domain adaptation and few-shot learning tasks. Specifically, our algorithm employs knowledge from these subgraphs to guide the adaptation process, enhancing our understanding of data distribution and improving the efficacy of domain transfer. Additionally, for the few-shot learning task, we implement a bi-level optimization strategy to determine how the extracted subgraph correlates with the labels, and adapt this understanding to various tasks.
We adopt a comprehensive strategy, integrating multiple techniques to systematically tackle each challenge. Methods for extracting core subgraph features focus on capturing critical features that remain stable across different tasks, while task-specific adaptation layers are designed to fine-tune the model’s parameters according to the specific requirements of the downstream tasks. This structured approach ensures our model's robustness and adaptability in handling diverse graph-based learning challenges.
Our contribution is summarized as follows:

\begin{itemize}[leftmargin=*]
\item  We introduce a novel framework, named \method{}, which learns the core subgraph instead of the whole graph representation learning. 
\item  We employ the learned core subgraph for the graph domain adaptation and few-shot learning task. By utilizing the core subgraph knowledge, the proposed \method{} enhances the robustness and scalability of graph classification.
\item  We show the effectiveness of our proposed \method{} with thorough experimentation, showing significant improvements in performance compared to leading methods across various datasets and evaluation metrics.
\end{itemize}

\begin{figure}[t]
\centering
\includegraphics[width=8.6cm,keepaspectratio=true]{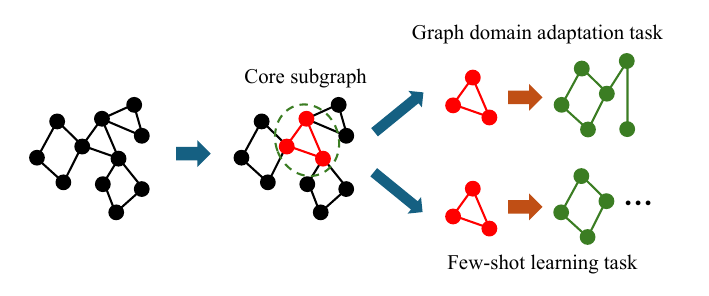}
\vspace{-0.4cm}
\caption{Illustration of the core knowledge learning framework. The framework extracts the core subgraph from the entire graph, which represents the fundamental structure necessary for task-relevant predictions. This core subgraph is then utilized for downstream tasks, including graph domain adaptation and few-shot learning tasks.}
\label{fig1}
\vspace{-0.2cm}
\end{figure}

\section{Related Work}\label{sec:related}

\textbf{Graph Classification}
Graph classification is a key task in graph-based machine learning, with wide applications in fields such as social network analysis, bioinformatics, and recommendation systems. The adoption of Graph Neural Networks (GNNs) has considerably pushed forward the discipline by facilitating the modeling of complex structures and relationships present in graph data \cite{kipf2017semisupervised}. These models excel in a variety of tasks including node classification, graph classification, and link prediction \cite{Rong2020DropEdge, hierarchical19, fan2019graph}. Despite their successes, traditional GNNs often face difficulties in effectively capturing higher-order topological structures like paths and motifs \cite{du2019graph}. In response, graph kernel methods have been developed to efficiently encapsulate structural information, offering a robust alternative to conventional GNN approaches \cite{long2021theoretically,yin2023omg}.

\textbf{Unsupervised Domain Adaptation}
Unsupervised domain adaptation (UDA) is a specialized area within machine learning focused on developing domain-invariant representations from a labeled source domain to be utilized in an unlabeled target domain. Traditional UDA methods typically involve aligning feature distributions between the source and target domains through techniques such as maximum mean discrepancy (MMD) \cite{long2015learning} or adversarial training \cite{ganin2016domain}.
Recent progress in UDA has been directed towards creating more efficient and scalable solutions to address domain shifts. A notable trend involves the adoption of deep learning strategies, including deep adversarial domain adaptation (DADA) \cite{tzeng2017adversarial} and domain-adaptive contrastive learning (DaCo) \cite{feng2023towards}. These approaches use neural networks to derive domain-invariant features. Additionally, methods like self-training \cite{yarowsky1995unsupervised} and pseudo-labeling \cite{lee2013pseudo} have been implemented to exploit unlabeled data in the target domain, enhancing the domain adaptation process.

\textbf{Graph Domain Adaptation}
Graph domain adaptation applies the concepts of unsupervised domain adaptation to graph-structured data. Its objective is to transfer insights from a labeled source graph to an unlabeled target graph, addressing challenges such as domain shift and label scarcity \cite{ju2024survey}. This issue is particularly pertinent in fields like social network analysis, where graphs depict social interactions, and bioinformatics, where graphs represent molecular structures. Existing methods mainly focus on how to transfer information from source graphs to unlabeled target graphs to learn effective node-level~\cite{wu2022attraction,zhu2021transfer,dai2022graph,guo2022learning} and graph-level~\cite{CoCo2023,yehudai2021local,ding2021closer,yang2020heterogeneous} representation.
Despite recent progress, the field continues to face obstacles such as misalignment in category distributions between source and target domains and the absence of scalable, effective algorithms for graph domain adaptation. Tackling these issues necessitates the development of robust, scalable algorithms capable of deriving domain-invariant representations from a limited amount of labeled data.

\textbf{Few-shot Learning}
Few-shot Learning (FSL) aims to train a model capable of generalizing to new classes based on only a small number of examples from those classes, often just one or a few. Meta-learning is a key strategy for FSL, enhancing the model's ability to generalize robustly. This technique involves extracting meta-knowledge that is applicable across a range of meta-tasks, enabling the model to adapt to new, unseen meta-tasks after sufficient meta-training. Meta-learning approaches for FSL can be divided into two main types: metric-based and optimization-based methods. Metric-based methods, such as Matching Network~\cite{vinyals2016matching} and ProtoNet~\cite{snell2017prototypical}, focus on learning a metric to assess similarity between new instances and a few examples by mapping them into a metric space. For example, Matching Network achieves this by encoding the support and query sets separately to calculate similarities, while ProtoNet creates prototypes by averaging the support set representations for each class and classifies queries based on the Euclidean distance to these prototypes. On the other hand, optimization-based methods~\cite{finn2017model,li2017meta,ravi2016optimization} concentrate on learning how to adjust model parameters efficiently using gradients from a few examples. MAML~\cite{finn2017model}, for instance, optimizes initial model parameters for quick fine-tuning with minimal examples. Another method employs an LSTM as a meta-learner to update parameters for specific tasks~\cite{ravi2016optimization}. Recently, FSL has been applied to graphs through meta-learning approaches~\cite{ding2020graph,liu2021relative}, demonstrating success in attributed networks. However, extracting meta-knowledge for satisfactory performance typically requires numerous meta-training tasks with a diverse class set and corresponding nodes. Our framework addresses the challenge of limited diversity in meta-tasks, offering a potential improvement over existing methods.

\section{Preliminary}


\subsection{Graph Neural Networks}

Considering the graph $G=(\mathcal{V}, \mathcal{E})$, let $\boldsymbol{h}_v^{(k)}$ represent the embedding vector of node $v$ at layer $k$. For each node $v \in \mathcal{V}$, we gather the embeddings of its neighbors from layer $k-1$. Subsequently, the embedding $\boldsymbol{h}_v^{(k)}$ is updated iteratively by merging $v$'s previous layer embedding with the embeddings aggregated from its neighbors. This process is formalized as follows:

\begin{equation}
\boldsymbol{h}_{v}^{(k)}= \mathcal{C}^{(k)}\left(\boldsymbol{h}_{v}^{(k-1)}, \mathcal{A}^{(k)} \left(\left\{\boldsymbol{h}_{u}^{(k-1)}\right\}_{u \in \mathcal{N}(v)}\right) \right),
\end{equation}
where $\mathcal{N}(v)$ represents the neighbors of $v$. $\mathcal{A}^{(k)}$ and $\mathcal{C}^{(k)}$ represent the aggregation and combination operations at the $k$-th layer, respectively. At last, we summarize all node representations at the $K$-th layer with a readout function into the graph-level representation, which can be formulated as follows:
\begin{equation}
\boldsymbol{z}= F\left(G\right)=\operatorname{READOUT}\left(\left\{\boldsymbol{h}_{v}^{(K)}\right\}_{v \in \mathcal{V}}\right),
\end{equation}
where $\boldsymbol{z}$ is the graph-level representation of $G$ and $\boldsymbol{\theta_e}$ denotes the parameter of our GNN-based encoder. $K$ denotes the number of the graph convolutional layers. The readout function can be implemented using different ways, such as the summarizing all nodes' representations~\cite{xu2019powerful} or using a virtual node~\cite{li2016gated}.  

After obtaining the graph representation $\boldsymbol{z}$, we introduce a multi-layer perception (MLP) classifier $H(\cdot)$ to output label distributions for final classification as follows:
\begin{equation}
    \hat{\textbf{p}} = H(\textbf{z}),
\end{equation}
where $ \hat{\textbf{p}} \in [0,1]^C $ and ${\boldsymbol{\theta_c}}$ denotes the parameters of the classifier.

\subsection{Explainer of GNN}
Following the methodology in~\cite{luo2020parameterized}, we partition the entire graph into two components, denoted as $G_{total}=G_{sub}+G_{rest}$. Here, $G_{sub}$ represents the critical subgraph that significantly influences the predictions of the GNN, and is thus considered the explanatory graph. Conversely, $G_{rest}$ includes the remaining edges that do not impact the GNN's predictions. To identify the essential subgraph $G_{sub}$, the approaches described in~\cite{luo2020parameterized, ying2019gnnexplainer} focus on maximizing the mutual information between the labels and $G_{sub}$:
\begin{equation}
\label{eq1}
    \max_{G_{sub}}MI(Y,G_{sub})=H(Y)-H(Y|G=G_{sub}),
\end{equation}
In this scenario, $Y$ represents the prediction made by the GNN when $G_{total}$ is used as input. Mutual information quantifies the likelihood of $Y$ when only a specific segment of the graph, $G_{sub}$, is processed by the GNN. This concept derives from conventional forward propagation methods used to provide clear explanations of how the model functions. For example, the importance of an edge $(i, j)$ is underscored when its removal leads to a significant change in the GNN’s output, suggesting that this edge is critical and should be included in $G_{sub}$. If an edge’s removal does not substantially affect the output, it is considered non-essential for the model's decision-making. Since $H(Y)$, the entropy of $Y$, is linked to the fixed parameters of the GNN during the explanation phase, the objective is to minimize the conditional entropy $H(Y|G = G_{sub})$.

Optimizing Eqn.~\ref{eq1} directly is impractical due to the $2^M$ potential candidates for $G_{sub}$, where $M$ represents the total number of edges. To simplify this, we assume the graph follows the Gilbert random graph model~\cite{gilbert1959random}, in which the edges of the subgraph are considered independent. Here, $e_{ij}=1$ indicates that the edge $(i,j)$ is included, and 0 indicates it is not. Under this model, the probability of any graph configuration can be expressed as a product of individual probabilities:
\begin{equation}
\label{eq2}
    P(G)=\Pi_{(i,j)\in\mathcal{V}}P(e_{ij}).
\end{equation}

Assuming the distribution of edge $e_{ij}$ follows the Bernoulli distribution: $e_{ij}\sim Bern(\theta_{ij})$. Then the Eqn.~\ref{eq2} can be rewrite as:
\begin{equation}
\begin{aligned}
        \min_{G_{sub}}H(Y|G=G_{sub})&=\min_{G_{sub}}\mathbb{E}_{G_{sub}}[H(Y|G=G_{sub})]\\&\approx \min_{\Theta}\mathbb{E}_{G_{sub}\sim q(\Theta)}[H(Y|G=G_{sub})],
\end{aligned}
\end{equation}
where $q(\Theta)$ is the distribution of the core subgraph.

\section{Methodology}

The purpose of learning core knowledge is to learn to determine the most essential subset of sample features. Focusing on the graph field, our purpose is to learn the subgraph structure that can represent the entire graph, and use the subgraph to replace the calculation of the entire graph. Motivated by~\cite{ma2019flexible}, where the real-life graphs are with underlying structures, we develop the core knowledge learning (CKL) module to learn the core subgraph of graphs and then utilize the core knowledge for the downstream tasks learning.

\begin{figure*}[t]
  \centering
  \includegraphics[scale=1]{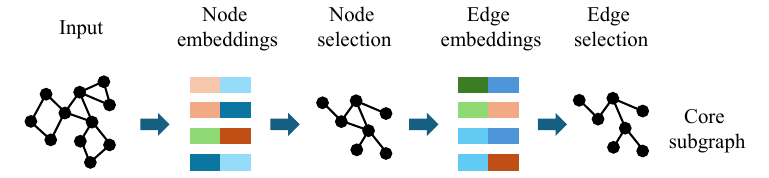}
  \caption{An overview of the proposed \method{}. \method{} first utilizes the node embeddings for node selection, and then cooperates the edge embeddings for edge selection to obtain the core subgraph.} 
  \label{fig1}
\end{figure*}

\subsection{Core Knowledge Learning}
To learn the core knowledge of a graph, we need to determine the important nodes subset from the whole graph. Traditional graph explainer methods~\cite{ma2019flexible,luo2020parameterized} direct learn the edges sampling process to determine the explainable subgraph. However, they typically assume the distribution of edges as prior knowledge, which is difficult to satisfy in real scenarios. Thus, we propose the CKL module to learn the probability of edge and node selection. Specifically, given a graph $G=(A,X)$, we first obtain the node embeddings with $l$-layer GNNs, and then map node $v_i$, $v_j$ and $e_{ij}$ into the same feature space with multilayer perception (MLP):
\begin{equation}
    H=GNN(A,X),\quad E_{n}=MLP(H),\quad E_{e}=MLP(E), 
\end{equation}
where $E$ denotes the features of edges.

\textbf{Node selection.} We first calculate the node sampling probability $p_v$ with the Sigmoid function to map the node embeddings into $[0,1]$:
\begin{equation}
    p_v=Sigmoid(E_{n_v}).
\end{equation}
A larger probability of node sampling will lead to a higher probability of node mask with $m_v=1$, indicating the corresponding node $v$ is important for the core knowledge learning. However, the node sampling process is non-differentiable~\cite{yu2023mind}, we relax $m_v$ with Gumbel-softmax~\cite{gal2017concrete,jang2016categorical}:
\begin{equation}
\label{node}
    m_v=Sigmoid(\frac{1}{t}\log \frac{p_v}{1-p_v}+\log\frac{u}{1-u}),
\end{equation}
where $t$ is the temperature parameter and $u\sim Uniform(0,1)$. 

\textbf{Edge selection.}  After obtaining the probability of node sampling, we further evaluate the probability of edges corresponding to the sampled nodes. Specifically, we concat the embeddings of edge $e_{ij}$ and the connected nodes $n_i$, $n_j$, and calculate the edge mask probability $m_{e_{ij}}$ with:
\begin{equation}
\label{edge}
    E_{fusion}=Cat(E_{n_i},E_{n_j},E_{e_{ij}}), \quad m_{e_{ij}}=Sigmod(E_{fusion}), 
\end{equation}
where $Cat$ denotes the concat operation.

With the node and edge selection process, we mask the whole graph $G_{total}$ to obtain the core graph $G_{sub}$. Finally, we follow~\cite{ying2019gnnexplainer} to modify the conditional entropy with cross-entropy $H(Y,Y_{sub})$, where $Y_{sub}$ is the prediction of the GNN model with $G_{sub}$ as the input:
\begin{equation}
\label{eq11}
    \min_{\Theta}\mathbb{E}_{G_{sub}\sim q(\Theta)} [H(Y|G=G_{sub})].
\end{equation}
where $q(\Theta)$ is the distribution of the core subgraph.

For efficient optimization of Eqn.~\ref{eq11}, we simplify the conditional entropy with cross-entropy $H(Y|Y_{sub})$, where $Y_{sub}$ is the output of subgraph $G_{sub}$. With the simplification, we optimize Eqn.~\ref{eq11} with Monte Carlo approximation:
\begin{equation}
\begin{aligned}
\label{eq12}
    \min_{\Theta} & \mathbb{E}_{G_{sub}\sim q(\Theta)} [H(Y|Y_{sub})]\\&\approx \min_{\Theta} -\frac{1}{K}\sum_{k=1}^K \sum_{c=1}^C P(Y=c)\log P(Y_{sub}=c)\\
    &=\min_{\Theta} -\frac{1}{K}\sum_{k=1}^K \sum_{c=1}^C P(Y=c|G=G_{total}) \log P(Y=c|G=G_{sub}^k),
\end{aligned}
\end{equation}
where $K$ is the number of sampled subgraphs, $C$ is the number of labels, and $G_{sub}^k$
 denotes the $k$-th sampled subgraph.

 \begin{figure}[t]
  \centering
  \includegraphics[scale=0.9]{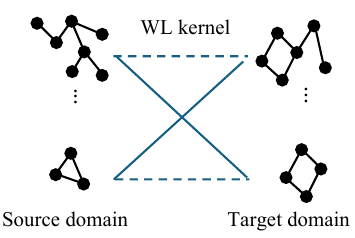}
  \caption{The core subgraph in the graph domain adaptation task. \method{} employs a kernel function to assess the similarity between source and target subgraphs, and assigning labels to the target graphs based on the most similar source graph. } 
  \vspace{-0.4cm}
  \label{fig1}
\end{figure}

\subsection{Graph Domain Adaptation Learning}
\textbf{Problem setup.}
Denote a graph as $G = (V, E,\mathbf{X})$ with the node set $V$, the edge set $E$, and the node attribute matrix $\mathbf{X} \in \mathbb{R}^{|V| \times F} $ with $F$ denotes the attribute dimension. The labeled source domain is denoted as $\mathcal{D}^s = \{(G_i^s, y_i^s)\}_{i=1}^{N_s}$, where $y_i^s$ denotes the labels of $G_i^s$. The unlabeled target domain is $\mathcal{D}^t = \{G_j^t\}_{j=1}^{N_t}$, where $N^s$ and $N^t$ denote the number of source graphs and target graphs. Both domains share the same label space $\mathcal{Y}$, but have different distributions in the graph space. Our objective is to train a model using both labeled source graphs and unlabeled target graphs to achieve superior performance in the target domain.

The extracted core subgraph is the underlying subgraph that makes important contribution to GNN's prediction and remaining is task-irrelevant part. Therefore, in the graph domain adaptation task, we measure the similarity of the source domain core subgraph and target domain core subgraph, ignoring the domain shift. Given two sampled subgraphs from source domain $G_{i}^S=(V_{i}^S,E_{i}^S)$ and target domain $G_{j}^T(V_{j}^T,E_{j}^T)$, graph kernels calculate their similarity by comparing their substructure using a kernel function. In formulation, 
\begin{equation}\label{eq:kernel}
K\left(G_{sub}^S, G_{sub}^T\right)=\sum_{v_{1} \in V_{i}^S} \sum_{v_{2} \in V_{j}^T} \kappa\left(l_{G_{sub}^S}\left(v_{1}\right), l_{G_{sub}^T}\left(v_{2}\right)\right)
\end{equation}
where $l_{G_{sub}^S}(v_1)$ represents the local substructure centered at node $v_1$ and $\kappa(\cdot, \cdot)$ is a pre-defined similarity measurement. We omit $l_G(\cdot)$ and leave $\kappa(u_1,u_2)$ in Eq. \ref{eq:kernel} for simplicity. In our implementation, we utilize the Weisfeiler-Lehmah (WL) subtree kernel for the comparison of source and target core subgraph.

\noindent\textbf{Weisfeiler-Lehmah (WL) Subtree Kernels.} WL subtree kernels compare all subtree patterns with limited depth rooted at every node. Given the maximum depth $l$, we have:
\begin{equation}
\begin{aligned} 
 K_{\text {subtree }}^{(i)}\left(G_{1}^S, G_{2}^T\right) &=\sum_{v_{1} \in V_{1}^S} \sum_{v_{2} \in V_{2}^T} \kappa_{\text {subtree }}^{(i)}\left(u_{1}, u_{2}\right) 
\\ K_{W L}\left(G_{1}^S, G_{2}^T\right) &=\sum_{i=0}^{l} K_{\text {subtree }}^{(i)}\left(G_{1}^S, G_{2}^T\right) \end{aligned}
\end{equation}
where $\kappa_{\text {subtree }}^{(i)}\left(u_{1}, u_{2}\right)$ is derived by counting matched subtree pairs of depth $i$ rooted at node $u_{1}$ and $u_{2}$, respectively. Considering the number of nodes of core subgraph is limited, we utilize the whole subgraph for the WL kernel calculation.


For each target domain core subgraph, we first calculate the most similar source domain subgraph and assign the same label to the target graph:
\begin{equation}
\label{eq15}
    Y_j^T=Y_i^S, \quad s.t. \max_{i,j} K(G_j^T, G_i^S).
\end{equation}
In this way, we avoid complex domain alignment operations and only align core subgraphs to achieve effective graph domain transfer learning.

 \begin{figure}[t]
  \centering
  \includegraphics[scale=0.9]{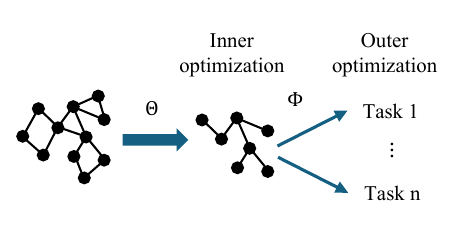}
  \vspace{-0.2cm}
  \caption{The core subgraph in the few-shot learning task. \method{} employs the bi-level method to optimize the core subgraph learning and multi-task prediction.} 
  \vspace{-0.4cm}
  \label{fig1}
\end{figure}

\subsection{Few shot learning}
\textbf{Problem setup.}  The target is to learn a predictor from a set of few-shot molecular property prediction tasks $\{\mathcal{T}_\tau\}_{\tau=1}^{N_t}$ and generalize to predict new properties given a few labeled molecules. The $\tau$-th task $\mathcal{T}_\tau$ predicts whether a molecule $x_{\tau,i}$ with index $i$ is active $(y_{\tau,i} = 1)$ or inactive $(y_{\tau,i} = 0)$ on a target property, provided with a small number of $K$ labeled samples per class. This $\mathcal{T}_\tau$ is then formulated as a 2-way K-shot classification task with a support set $\mathcal{S}_\tau=\{(x_{\tau,i},y_{\tau,i})\}_{i=1}^{2K}$ containing the $2K$ labeled samples and a query set $\mathcal{Q}_\tau=\{(x_{\tau,j},y_{\tau,j}\}_{j=1}^{N_\tau^q}$ containing $N_\tau^q$ unlabeled samples to be classified.

In order to achieve the goal of few shot learning on multiple tasks, we utilize the learned core subgraph as the input of GNN for prediction, i.e.,
\begin{equation}
    \hat{Y}=GNN_\Phi\left(G_{sub}(\Theta)\right),
\end{equation}
where $\Phi$ is the parameters of task relevant embedding function and classifier and $\Theta$ denotes the collection of parameters of CKL molecular. The training loss $\mathcal{L}(\mathcal{S}_{\tau}, f_{\theta,\Phi})$ evaluated on $\mathcal{S}_{\tau}$ follows:
\begin{equation}
\label{eq17}
    \mathcal{L}(\mathcal{S}_\tau,f_{\Theta,\Phi})=\sum_{(x_{\tau,i}y_{\tau,i})\in\mathcal{S}_\tau} -\mathbf{y}_{\tau,i}^\top log(\hat{\mathbf{y}}_{\tau,i}),
\end{equation}
where $\mathbf{y}_{\tau,i}\in\mathbb{R}^2$ is a one-hot ground-truth.
Observing that the Eqn.~\ref{eq17} contains two distinct parameters $\Theta$ and $\Phi$, we further use bi-level optimization methods to optimize them simultaneously:

\begin{equation}
\begin{aligned}
    \min_{\Phi} F(\Theta^\star)=\sum\mathcal{L}_{outer}\left(f_{\Theta^\star,\Phi}(A,X,\mathcal{S}_\tau)\right),\\
    s.t. \Theta^\star=\sum\mathcal{L}_{inner}\left(f_{\Theta,\Phi}(A,X,\mathcal{S}_\tau)\right),
\end{aligned}
\end{equation}
where $\mathcal{L}_{inner}$ is the loss function in Eqn.~\ref{eq12} and $\mathcal{L}_{outer}$ is the loss function of Eqn,~\ref{eq17}. 

\begin{algorithm}[t]
\caption{Learning Algorithm of \method{} }
\label{alg}
\begin{flushleft}
\textbf{Input:} Training graphs, test graphs. \\
\textbf{Output}: Parameters $\Theta$ and ${\Phi}$ for different tasks.  \\
\end{flushleft}
\begin{algorithmic}[1] 
\STATE \textrm{// Core subgraph learning}\\
\STATE For each training and test data, select core edges and nodes with Eqn.~\ref{node} and \ref{edge};
\STATE Optimize the core subgraph learning parameter $\Theta$ with Eqn.~\ref{eq12};
\\
\STATE
\STATE \textrm{// Graph domain adaptation task}\\
\FOR{$G_i^T$ in $G_{sub}^T$}
\STATE Calculate the similarity of $G_i^T$ with the source domain subgraph $G_{sug}^S$;
\STATE Assign the label to $G_i^T$ with Eqn.~\ref{eq15};
\ENDFOR
\\
\STATE
\STATE \textrm{// Few-shot learning task}\\
\WHILE{not early stopping}
\STATE \textrm{// inner optimization}\\
\FOR{$t=1$ to $T$}  
\STATE Update $\Theta$ with Eqn.~\ref{eq19};  
\ENDFOR
\FOR{$t=T$ downto 1}
\STATE Calculate the outer gradient with Eqn. 4 in~\cite{yin2022generic};
\ENDFOR
\STATE \textrm{// outer optimization}\\
\STATE Update $\Phi$ with Eqn.~\ref{eq21};
\ENDWHILE
\end{algorithmic}
\end{algorithm}

\textbf{Inner optimization.} We first optimize the parameter $\Phi$ with a gradient descent based optimizer by fixing $\Theta$, 
\begin{equation}
\label{eq19}
    \Theta_{t}=\Theta_{t-1}-\alpha \nabla_{\Theta} \mathcal{L}_{inner}(\mathcal{S}_\tau,f_{\Theta,\Phi}),
\end{equation}
where $\alpha$ is the learning rate.

\textbf{Outer optimization.} Following~\cite{finn2017model}, we employ the gradient-based meta-learning strategy and initialize $\Phi$ with set of meta-training tasks $\{\mathcal{T}_{\tau} \}^{N_t}_{\tau=1}$, which acts as the anchor of each task $\mathcal{T}_{\tau}$. Specifically, we fix parameter $\Theta$ and optimize $\Phi$ as $\Phi_\tau$ on $\mathcal{S}_\tau$ in each $\mathcal{T}_{\tau}$ during the outer optimization period. 
 $\Phi_\tau$ is obtained by taking a few gradient descent updates:
\begin{equation}
    \nabla_\Phi \mathcal{L}(\mathcal{S}_\tau,f_{\Theta,\Phi})=\partial_{\Theta}\mathcal{L}_{outer} \nabla_\Phi \Theta(\Phi)+\partial_\Phi \mathcal{L}_{outer}.
\end{equation}
We follow~\cite{yin2022generic} to optimize the parameter $\Phi$ with:
\begin{equation}
\label{eq21}
    \Phi=\Phi-\beta \nabla_{\Phi}\mathcal{L}_{outer}(f_{\Theta^\star,\Phi}(A,X,\mathcal{S}_\tau)).
\end{equation}

The details of updating $\Theta$ and $\Phi$ are shown in Algorithm 1. The core knowledge learning is shown in lines 2-3, the graph domain adaptation task is shown in lines 6-9, and the few-shot learning is shown in lines 12-22. 



\section{Experiments}
\subsection{Experimental Settings}

\begin{table}[t]
\centering
\tabcolsep=2pt
\caption{Introduction of datasets on domain adaptation task.}
\begin{tabular}{llcccc}
\toprule
\multicolumn{2}{c}{Datasets}  &Graphs  &Avg. Nodes   &Avg. Edges &Classes \\
\midrule 
\multicolumn{2}{l}{Mutagenicity} &4337 &30.32 &30.77 &2\\
\midrule
\multicolumn{2}{l}{Tox21\_AhR} &8169 &18.09 &18.50 &2\\
\midrule
\multicolumn{2}{l}{FRANKENSTEIN} &4337 &16.9 &17.88 &2\\
\midrule 
\multirow{2}*{PROTEINS} &PROTEINS &1,113 &39.1 &72.8 &2 \\ &DD &1,178 &284.3 &715.7 &2 \\
\midrule 
\multirow{2}*{COX2} &COX2 &467 &41.2 &43.5 &2 \\ &COX2\_MD &303 &26.3 &335.1 &2\\
\midrule 
\multirow{2}*{BZR} &BZR &405 &35.8 &38.4 &2 \\ &BZR\_MD &306 &21.3 &225.1 &2 \\
\bottomrule
\end{tabular}

\label{GDA_task}
\end{table}

\begin{table}[t]
\centering
\tabcolsep=7pt
\caption{Introduction of datasets on few-shot learning task.}
\begin{tabular}{lcccc}
\toprule
{\bf Dataset} &Tox21 &SIDER &MUV &ToxCast\\
\midrule
Compounds &8014 &1427 &93127 &8615\\
Tasks &12 &27 &17 &617\\
Training Tasks &9 &21 &12 &450\\
Testing Tasks &3 &6 &5 &167\\
\bottomrule
\end{tabular}
\label{few-shot}
\vspace{-0.2cm}
\end{table}

\noindent\textbf{Datasets.}
For the graph domain adaptation task, we utilize 9 graph classification datasets for evaluation, i.e., Mutagenicity (M)~\cite{kazius2005derivation}, Tox21\_AhR~\footnote{https://tripod.nih.gov/tox21/challenge/data.jsp}, FRANKENSTEIN (F)~\cite{orsini2015graph}, and PROTEINS~\cite{dobson2003distinguishing} (including PROTEINS (P) and DD (D)), COX2~\cite{sutherland2003spline} (including COX2 (C) and COX2\_MD (CM)), BZR~\cite{sutherland2003spline} (including BZR (B) and BZR\_MD (BM)) obtained from TUDataset~\cite{Morris2020}. The details statistics are presented in Table~\ref{GDA_task}. Additionally, we follow~\cite{CoCo2023} and partition M, T, F datasets into four sub-datasets based on edge density. For the few-shot learning task, we evaluate the experiments on widely used few-shot molecular property prediction, and the details of the datasets are introduced in Table~\ref{few-shot}.

\noindent\textbf{Baselines.}
For the graph domain adaptation task, we compare our \method{} with different baselines: {WL subtree}~\cite{shervashidze2011weisfeiler} {GCN}~\cite{kipf2017semi}, {GIN}~\cite{xu2019powerful}, {GMT}~\cite{BaekKH21}, {CIN}~\cite{bodnar2021weisfeiler}, {CDAN}~\cite{long2018conditional}, {ToAlign}~\cite{NeurIPS2021_731c83db}, {MetaAlign}~\cite{wei2021metaalign}, DEAL~\cite{yin2022deal} and CoCo~\cite{CoCo2023}. For the few-shot learning task, we compare \method{} with Siamese~\cite{koch2015siamese}, ProtoNet~\cite{snell2017prototypical}, MAML~\cite{finn2017model}, TPN~\cite{liu2018learning}, EGNN~\cite{kim2019edge}, IterRefLSTM~\cite{altae2017low} and RAP~\cite{wang2021property}.

\noindent\textbf{Implementation Details.}
\label{setting}
In our \method{}, we employ {GIN}~\cite{xu2019powerful} as the backbone of feature extraction. For the graph domain adaptation task, we utilize one of the sub-datasets as source data and the remaining as the target data for performance comparison. We set the hidden size to 128 and the learning rate to 0.001 as default. We report the classification accuracy in the experiments. For the few-shot learning task, we use RDKit~\cite{landrum2013rdkit} to obtain the molecular graphs, node and edge features. We use the GIN~\cite{xu2019powerful} as the backbone for feature extraction. We calculate the mean and standard deviations of ROC-AUC scores on each task by running ten times experiments.

\begin{table}[t]
\centering
\tabcolsep=1.8pt
\caption{The classification results (in \%) on PROTEINS, COX2, and BZR domain shift (source$\rightarrow$target). P, D denote the PROTEINS, DD, C and CM denote the COX2 and COZ2\_MD, B and BM denote BZR and BZR\_MD. \textbf{Bold} results indicate the best performance.}
\begin{tabular}{l|c|c|c|c|c|c|c}
\toprule
{\bf Methods} &P$\rightarrow$D &D$\rightarrow$P &C$\rightarrow$CM &CM$\rightarrow$C &B$\rightarrow$BM &BM$\rightarrow$B &Avg.\\
\midrule
WL subtree  &72.9 &41.1 &48.8 &78.2 &51.3 &78.8 &61.9\\
GCN &58.7  &59.6  &51.1 &78.2 &51.3 &71.2 &61.7\\
GIN &61.3 &56.8 &51.2 &78.2 &48.7 &78.8 &62.5\\
CIN &62.1 &59.7 &57.4 &61.5 &54.2 &72.6 &61.3\\
GMT &62.7 &59.6 &51.2 &72.2 &52.8 &71.3 &61.6\\
CDAN &59.7 &64.5 &59.4 &78.2 &57.2 &78.8 &66.3\\
ToAlign &62.6 &64.7 &51.2 &78.2 &58.4 &78.7 &65.7\\
MetaAlign &60.3 &64.7 &51.0 &77.5 &53.6 &78.5 &64.3\\
\midrule 
DEAL  &{76.2} &63.6 &{62.0} &{78.2}  &{58.5} &{78.8} &{69.6}\\
 CoCo  &74.6 &\textbf{67.0} &61.1 &{79.0} &{62.7} &{78.8} &{70.5}\\
 \midrule
 \method{} &\textbf{76.8} &66.4 &\textbf{62.8} &\textbf{79.3} &\textbf{62.8} &\textbf{79.0} &\textbf{71.2}\\
\bottomrule
\end{tabular}
\label{table3}
\vspace{-0.4cm}
\end{table}

\subsection{Performance on Different Domains}

\begin{table*}[t]
\centering
\tabcolsep=2pt
\caption{The classification results (in \%) on Mutagenicity under edge density domain shift (source$\rightarrow$target). M0, M1, M2, and M3 denote the sub-datasets partitioned with edge density. \textbf{Bold} results indicate the best performance.}
\vspace{-0.2cm}
\resizebox{\textwidth}{!}{
\begin{tabular}{l|c|c|c|c|c|c|c|c|c|c|c|c|c}
\toprule
{\bf Methods} &M0$\rightarrow$M1 &M1$\rightarrow$M0 &M0$\rightarrow$M2 &M2$\rightarrow$M0 &M0$\rightarrow$M3 &M3$\rightarrow$M0 &M1$\rightarrow$M2 &M2$\rightarrow$M1 &M1$\rightarrow$M3 &M3$\rightarrow$M1 &M2$\rightarrow$M3 &M3$\rightarrow$M2 &Avg.\\
\midrule
WL subtree  &74.9 &74.8 &67.3 &69.9 &57.8 &57.9 &73.7 &80.2 &60.0 &57.9 &70.2 &73.1 &68.1\\
GCN &71.1 &70.4 &62.7 &69.0 &57.7 &59.6 &68.8 &74.2 &53.6 &63.3 &65.8 &74.5 &65.9\\
GIN &72.3 &68.5 &64.1 &72.1 &56.6 &61.1 &67.4 &74.4 &55.9 &67.3 &62.8 &73.0 &66.3\\
CIN &66.8 &69.4 &66.8 &60.5 &53.5 &54.2 &57.8 &69.8 &55.3 &74.0 &58.9 &59.5 &62.2\\
GMT &73.6 &75.8 &65.6 &73.0 &56.7 &54.4 &72.8 &77.8 &62.0 &50.6 &64.0 &63.3 &65.8\\
CDAN &73.8 &74.1 &68.9 &71.4 &57.9 &59.6 &70.0 &74.1 &60.4 &67.1 &59.2 &63.6 &66.7\\
ToAlign &74.0 &72.7 &69.1 &65.2 &54.7 &73.1 &71.7 &77.2 &58.7 &73.1 &61.5 &62.2 &67.8\\
MetaAlign &66.7 &51.4 &57.0 &51.4 &46.4 &51.4 &57.0 &66.7 &46.4 &66.7 &46.4 &57.0 &55.4\\
\midrule 
DEAL  &77.5 &75.7  &68.3 &74.9 &65.1 &74.0 &76.9 &77.4 &66.4 &71.2 &62.8 &77.1 &72.2\\
CoCo  &{77.7} &{76.6}  &{73.3} &{74.5} &{66.6} &{74.3} &77.3 &{80.8} &{67.4} &{74.1} &68.9 &{77.5} &{74.1}\\
\midrule 
\method{} & \textbf{78.6} & \textbf{76.8} & \textbf{73.9} & \textbf{75.4} & \textbf{68.2} &\textbf{75.3} & \textbf{78.5} & \textbf{81.3} & \textbf{67.9} & \textbf{75.2} & \textbf{69.4} & \textbf{78.4} & \textbf{74.9}\\
\bottomrule
\end{tabular}}
\label{table4}
\vspace{-0.2cm}
\end{table*}

\begin{table*}[t]
\centering
\tabcolsep=4.5pt
\caption{The classification results (in \%) on Tox21 under edge density domain shift (source$\rightarrow$target). T0, T1, T2, and T3 denote the sub-datasets partitioned with edge density. \textbf{Bold} results indicate the best performance.}
\vspace{-0.2cm}
    \resizebox{\textwidth}{!}{
\begin{tabular}{l|c|c|c|c|c|c|c|c|c|c|c|c|c}
\toprule
{\bf Methods} &T0$\rightarrow$T1 &T1$\rightarrow$T0 &T0$\rightarrow$T2 &T2$\rightarrow$T0 &T0$\rightarrow$T3 &T3$\rightarrow$T0 &T1$\rightarrow$T2 &T2$\rightarrow$T1 &T1$\rightarrow$T3 &T3$\rightarrow$T1 &T2$\rightarrow$T3 &T3$\rightarrow$T2 &Avg.\\
\midrule
WL subtree  &65.3 &51.1 &69.6 &52.8 &53.1 &54.4 &71.8 &65.4 &60.3 &61.9 &57.4 &76.3 &61.6\\
GCN &64.2 &50.3 &67.9 &50.4 &52.2 &53.8 &68.7 &61.9 &59.2 &51.4 &54.9 &76.3 &59.3\\
GIN &67.8 &51.0 &77.5 &54.3 &56.8 &54.5 &78.3 &63.7 &56.8 &53.3 &56.8 &77.1 &62.3\\
CIN &67.8 &50.3 &78.3 &54.5 &56.8 &54.5 &78.3 &67.8 &59.0 &67.8 &56.8 &78.3 &64.2\\
GMT &67.8 &50.0 &78.4 &50.1 &56.8 &50.7 &78.3 &67.8 &56.8 &67.8 &56.4 &78.1 &63.3\\
CDAN &69.9 &55.2 &78.3 &56.0 &59.5 &56.6 &78.3 &\textbf{68.5} &61.7 &68.1 &61.0 &78.3 &66.0\\
ToAlign &68.2 &58.5 &78.4 &58.8 &58.5 &53.8 &{78.8} &67.1 &64.4 &68.8 &57.9 &78.4 &66.0\\
MetaAlign &65.7 &57.5 &78.0 &58.5 &\textbf{63.9} &52.2 &78.8 &67.1 &62.3 &67.5 &56.8 &78.4 &65.6\\

\midrule 
DEAL &68.8 &54.7 &76.3 &56.9 &61.8 &57.8 &77.4 &65.6 &63.7 &67.2 &60.4 &77.9 &65.7\\
CoCo   &{69.9} &{59.8} &{78.8}  &{59.0} &62.3 &{59.0} &78.4 &66.8 &{65.0} &{68.8} &{61.2} &{78.4} &{67.3}\\
\midrule
\method{}  &\textbf{70.2} &\textbf{60.3} &\textbf{79.5} &\textbf{59.3} &63.7 &\textbf{59.8} &\textbf{79.2} &67.7 &\textbf{65.7} &\textbf{69.4} &\textbf{61.7} &\textbf{78.9} &\textbf{68.0}\\
\bottomrule
\end{tabular}}
\label{table5}
\vspace{-0.2cm}
\end{table*}

\begin{figure*}[h]
  \centering
  \includegraphics[scale=0.58]{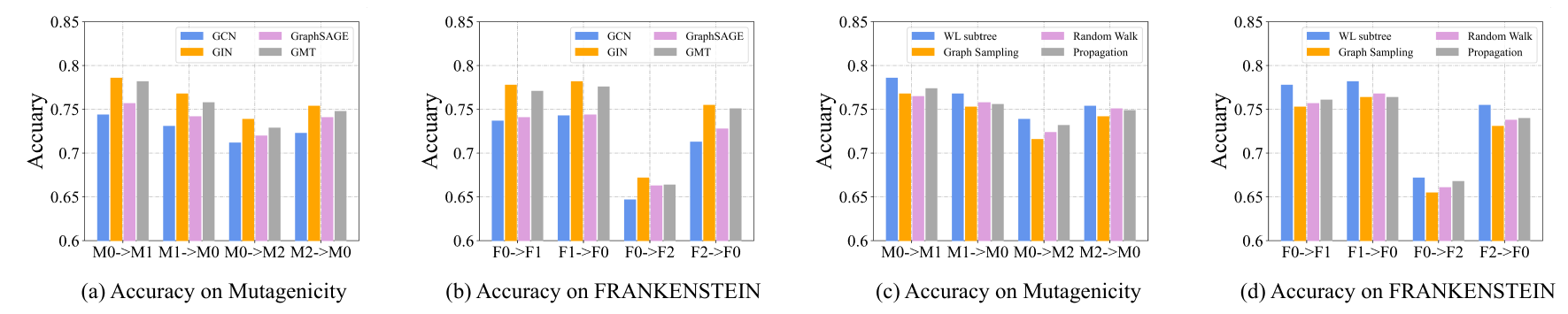}
\vspace{-0.2cm}  
\caption{The performance with different GNNs and kernels on different datasets. (a), (b) are the performance of different GNNs, (c), (d) are the performance of different graph kernels.}
  \label{GNN}
  \vspace{-0.2cm}
\end{figure*}

Tables \ref{table3} to \ref{table6} present the comparative results of \method{} alongside other benchmark methods. Analyzing these results, we observe several key insights:

\begin{itemize}[leftmargin=*]
\item Superiority of Domain Adaptation Methods in Graphs: Domain adaptation strategies tailored for graphs consistently outperform traditional kernel and GNN-based methods. This suggests that conventional graph methodologies may struggle with adaptability across varying domains due to their limited expressive power. Therefore, the development of domain-invariant techniques is critical for the advancement of Graph Domain Adaptation (GDA). These domain-invariant methods prove essential not only in maintaining performance across diverse datasets but also in facilitating the integration of graphs from disparate sources without loss of fidelity.

\item Robust Performance of GDA Techniques: Methods implemented in GDA demonstrate robust performance, notably surpassing traditional domain adaptation strategies. The success of these methods can be attributed to their ability to handle the inherent complexities in graph data. Achieving high-quality graph representations is a complex task, exacerbated by the structural and feature diversity within the graphs. This complexity renders traditional domain adaptation strategies less effective, thus highlighting the specialized nature and effectiveness of graph-specific adaptation methods.

\item Efficiency of \method{}: The proposed \method{} method outshines other competing methods, showcasing its efficiency in core knowledge learning. This efficiency is largely due to \method{}'s focus on critical components essential for accurate predictions. Specifically, \method{} excels by concentrating on learning and enhancing the most relevant parts of the graph for the task at hand, while effectively ignoring or minimizing attention to the less relevant subgraphs. This approach not only boosts performance but also enhances the model's ability to generalize across different tasks by reducing the noise associated with irrelevant data.

\end{itemize}


\begin{table*}[t]
\centering
\tabcolsep=4.5pt
\caption{Graph classification accuracy (in \%) on FRANKENSTEIN under edge density domain shift (source$\rightarrow$target). F0, F1, F2, and F3 denote the sub-datasets partitioned with edge density. \textbf{Bold} results indicate the best performance.}
\begin{tabular}{l|c|c|c|c|c|c|c|c|c|c|c|c|c}
\toprule
{\bf Methods} &F0$\rightarrow$F1 &F1$\rightarrow$F0 &F0$\rightarrow$F2 &F2$\rightarrow$F0 &F0$\rightarrow$F3 &F3$\rightarrow$F0 &F1$\rightarrow$F2 &F2$\rightarrow$F1 &F1$\rightarrow$F3 &F3$\rightarrow$F1 &F2$\rightarrow$F3 &F3$\rightarrow$F2 &Avg.\\
\midrule
WL subtree  &71.6 &72.1 &62.1 &71.2 &57.8 &67.7 &64.0 &75.3 &41.1 &59.2 &55.9 &55.4 &62.8\\
GCN &75.7  &74.1 &65.7 &73.9 &52.5 &72.7 &65.4 &73.7 &54.1 &58.3 &60.9 &57.0 &65.3\\
GIN &76.7 &{76.9} &63.3 &72.4 &55.5 &53.6 &65.1 &75.4 &58.2 &66.1 &55.9 &58.7 &64.8\\
CIN &76.6 &76.5 &63.4 &68.8 &57.6 &71.7 &64.3 &72.0 &50.2 &70.2 &54.5 &55.7 &65.1\\
GMT &67.2 &64.0 &55.4 &51.1 &60.4 &58.8 &62.5 &64.2 &60.4 &50.6 &60.2 &57.9 &59.4\\
CDAN &72.7 &74.1 &63.1 &73.7 &60.1 &68.7 &62.7 &72.6 &60.4 &70.2 &60.5 &59.4 &66.5\\
ToAlign &72.5 &76.6 &64.4 &67.1 &\textbf{60.7} &63.7 &65.0 &74.9 &59.5 &67.9 &61.0 &57.8 &65.9\\
MetaAlign &73.5 &70.0 &64.6 &60.2 &60.4 &62.8 &64.6 &71.5 &60.4 &63.5 &60.4 &54.6 &63.9\\

\midrule
DEAL   &77.4 &75.7 &{66.7} &{74.2}  &58.2 &\textbf{73.7} &{65.5} &{75.6} &{61.1} &{70.6} &{61.7} &{59.7} &{68.3}\\
CoCo &74.6 &77.2 &64.1 &73.8 &60.5 &71.5 &65.9 &76.0 &61.4 &72.6 &59.6 &64.7 &68.5\\
\midrule 
\method{}  &\textbf{77.8} & \textbf{78.2} & \textbf{67.2} & \textbf{75.5} & \textbf{62.2} & {72.6} & \textbf{68.4} & \textbf{77.9} & \textbf{62.7} & \textbf{73.3} & \textbf{62.1} & \textbf{66.3} & \textbf{70.4}\\
\bottomrule
\end{tabular}
\label{table6}
\end{table*}

\begin{table*}[t]
\centering
\tabcolsep=9pt
\caption{ROC-AUC scores on benchmark molecular property prediction datasets. Bold results (according to the pairwise t-test with 95\% confidence) indicate the best performance.}
\begin{tabular}{l|c|c|c|c|c|c|c|c}
\toprule
\multirow{2}{*}{Methods} & \multicolumn{2}{c|}{Tox21} &\multicolumn{2}{c|}{SIDER} &\multicolumn{2}{c|}{MUV} &\multicolumn{2}{c}{ToxCast} \\
\cline{2-9} &10-shot &1-shot &10-shot &1-shot &10-shot &1-shot &10-shot &1-shot\\
\midrule
Siamese &80.4$\pm$0.4 &65.0$\pm$1.6 &71.1$\pm$4.3 &51.4$\pm$3.3 &60.0$\pm$5.1 &50.0$\pm$0.2 &- &-\\
ProtoNet &75.0$\pm$0.3 &65.6$\pm$1.7 &64.5$\pm$0.9 &57.5$\pm$2.3 &65.9$\pm$4.1 &58.3$\pm$3.2 &63.7$\pm$1.3 &56.4$\pm$1.5\\
MAML &80.2$\pm$0.2 &75.7$\pm$0.5 &70.4$\pm$0.8 &67.8$\pm$1.1 &63.9$\pm$2.3 &60.5$\pm$3.1 &66.8$\pm$0.9 &66.0$\pm$5.0\\
TPN &76.1$\pm$0.2 &60.2$\pm$1.2 &67.8$\pm$1.0 &62.9$\pm$1.4 &65.2$\pm$5.8 &50.0$\pm$0.5 &62.7$\pm$1.5 &50.0$\pm$0.1\\
EGNN &81.2$\pm$0.2 &79.4$\pm$0.2 &72.9$\pm$0.7 &70.8$\pm$1.0 &65.2$\pm$2.1 &62.2$\pm$1.8 &63.7$\pm$1.6 &61.0$\pm$1.9\\
IterRefLSTM &81.1$\pm$0.2 &81.0$\pm$0.1 &69.6$\pm$0.3 &71.7$\pm$0.1 &49.6$\pm$5.1 &48.5$\pm$3.1 &- &-\\
PAR &82.1$\pm$0.1 &80.5$\pm$0.1 &74.7$\pm$0.3 &\textbf{71.9$\pm$0.5} &66.5$\pm$2.1 &64.1$\pm$1.2 &69.7$\pm$1.6 &67.3$\pm$2.9\\
\midrule 
\method{}  &\textbf{82.3$\pm$0.3} & \textbf{81.4$\pm$0.4} & \textbf{75.3$\pm$0.5} & {71.7$\pm$0.7} & \textbf{67.1$\pm$2.3} & \textbf{64.4$\pm$2.2} & \textbf{70.2$\pm$1.5} & \textbf{68.1$\pm$1.6} \\
\bottomrule
\end{tabular}
\vspace{-0.4cm}
\label{table7}
\end{table*}

\vspace{-0.1cm}
\subsection{Performance on Few-shot Learning}


Table~\ref{table7} details the performance comparisons between \method{} and a range of baseline methods in graph-based molecular encoding tasks. For this analysis, we have omitted results pertaining to Siamese and IterRefLSTM, as their implementation details and the outcomes of their evaluations on the ToxCast dataset are unavailable. The table clearly illustrates that \method{} consistently achieves superior performance over other methods that also employ graph-based molecular encoders designed from the ground up. In particular, \method{} not only surpasses all compared baselines but does so with a notable margin; it demonstrates an average performance improvement of 1.62\% over the highest-performing baseline, EGNN.
The performance reveals that methods incorporating few-shot learning techniques to decode relation graphs, such as GNN, TPN, and EGNN, deliver enhanced results when contrasted with more traditional learning frameworks like ProtoNet and MAML. This observation underscores the efficacy of few-shot learning in complex graph analysis tasks, where the ability to rapidly adapt to new, limited data without extensive retraining provides a significant advantage. These few-shot learning methods leverage their sophisticated algorithms to effectively capture and utilize the intricate relationships and structural nuances present within the graph data, thereby yielding more accurate and reliable predictions.

\vspace{-0.2cm}
\subsection{Flexibility of \method{}}
For the graph domain adaptation experiments, we use GIN as the backbone to extract the core subgraph feature. To show the flexibility of the proposed \method{}, we replace the GIN with different GNN methods. In our implementation, we utilize GCN~\cite{kipf2017semi}, GraphSage~\cite{hamilton2017inductive} and GMT~\cite{BaekKH21} instead of GIN to show the flexibility of \method{}. Additionally, we replace the WL subtree kernel with Graph Sampling~\cite{leskovec2006sampling}, Random Walk~\cite{kalofolias2021susan} and Propagation~\cite{neumann2016propagation}.

Figure~\ref{GNN} illustrates the comparative performance of several GNNs and graph kernels over four representative datasets. We have noted similar performance trends across additional datasets as well. The data indicate that among the various GNNs and graph kernels evaluated, GIN and the WL subtree kernel consistently stand out as the top performers in the majority of cases. The superior performance of both GIN and the WL subtree kernel is likely due to their exceptional capabilities in capturing complex graph structures and providing powerful node and graph-level representations. This consistent outperformance validates our selection of GIN and the WL subtree kernel as the primary methods for enhancing task performance in our graph domain adaptation efforts. The choice is further justified by their ability to effectively handle the complexities of diverse datasets, making them highly suitable for robust graph analysis and domain adaptation tasks.

\section{Conclusion}
In this paper, we introduce a novel approach named \method{} that focuses on learning the core subgraph knowledge necessary for downstream tasks. Recognizing the essential role of the underlying subgraph in GNN predictions, while considering the rest as task-irrelevant, we have developed a framework designed for graph adaptation and scalability learning. \method{} includes several key components: the core subgraph knowledge submodule, the graph domain adaptation module, and the few-shot learning module, each aimed at addressing specific challenges in graph classification such as domain shifts, label inconsistencies, and data scarcity. Our comprehensive experiments show that \method{} significantly outperforms existing state-of-the-art methods, demonstrating notable advancements in performance.

\clearpage
\bibliographystyle{ACM-Reference-Format}
\bibliography{sample-base}

\end{document}